\documentclass[runningheads]{llncs}
\usepackage{rotating}
\usepackage{graphicx}
\usepackage{array}
\usepackage{mdwmath}
\usepackage{mdwtab}
\usepackage{eqparbox}
\usepackage{url}
\usepackage{threeparttable}
\usepackage[abs]{overpic}

\usepackage{soul}
\usepackage{multirow}
\usepackage{verbatim}
\usepackage{bbm}
\usepackage{color}
\newcommand{\tabincell}[2]{\begin{tabular}{@{}#1@{}}#2\end{tabular}}
\begin{document}
\title{Cephalometric Landmark Detection by Attentive Feature Pyramid Fusion and Regression-Voting}
\titlerunning{Cephalometric Landmark Detection by AFPF and Regression-Voting}
%

\author{Runnan Chen\inst{1} \and
Yuexin Ma\inst{1} \and
Nenglun Chen\inst{1}\and
Daniel Lee\inst{2}\and
Wenping Wang\inst{1}}
\authorrunning{R. Chen et al.}

\institute{Department of Computer Science, The University of Hong Kong, Hong Kong \\ \and
Modontics (Hong Kong) Limited, Hong Kong\\
}

\maketitle              

\begin{abstract}
Marking anatomical landmarks in cephalometric radiography is a critical operation in cephalometric analysis. Automatically and accurately locating these landmarks is a challenging issue because different landmarks require different levels of resolutions and semantics. Based on this observation, we propose a novel attentive feature pyramid fusion module (AFPF) to explicitly shape high-resolution and semantically enhanced fusion features to achieve significantly higher accuracy than existing deep learning-based methods. We also combine heat maps and offset maps to perform pixel-wise regression-voting to improve  detection accuracy. By incorporating the AFPF and regression-voting, we develop an end-to-end deep learning framework that improves detection accuracy by \textbf{7\%$\sim$11\%} for all the evaluation metrics over the state-of-the-art method. We present ablation studies to give more insights into different components of our method and demonstrate its generalization capability and stability for unseen data from diverse devices.

\keywords{cephalometric landmarks, deep learning, self-attention, fusion feature, regression-voting.}
\end{abstract}
\section{Introduction}
Cephalometric analysis is widely used in evaluation and treatment planning for orthodontic, orthognathic and maxillofacial surgeries. It provides the clinician with crucial information on the patient's dental, skeletal and facial relationship. The key operation during the analysis is marking craniofacial landmarks~\cite{ricketts1982orthodontic} to assess and quantify the degree of the anatomical abnormalities. In practice, landmarks are located manually, which is tedious, time-consuming, and unreliable in achieving reproducible results. Hence, fully automatic and accurate landmark localization has been a long-standing area with a great deal of need.

Current solutions can be classified into five categories:  knowledge-based, pattern matching-based, statistical learning-based, hybrid-based and deep learning-based methods. The first category~\cite{levy1986knowledge} is to simulate the manual landmark detection process with human knowledge of the landmark structures. However, rules become too complex to be formulated with the increase of the image complexity. Then, some researchers employed search methods using pattern matching~\cite{cardillo1994image,el2004automatic}. However, they are quite sensitive to individual variations. Considering that both the global spatial constraints and local appearance of landmark locations are important, some statistical learning-based approaches have been proposed, like Active Shape Model~\cite{cootes1995active} and Active Appearance Model~\cite{cootes2001active}. Two frameworks~\cite{ibragimov2014automatic,lindner2015fully} combining the random forests regression-voting and the statistical shape analysis techniques perform well in the IEEE ISBI 2014 and 2015 Challenges~\cite{wang2015evaluation,wang2016benchmark}. Since then, almost all the methods are benchmarked against the Grand Challenges dataset~\cite{lindner2016fully,arik2017fully,payer2019integrating,urschler2018integrating}. There are also some hybrid-based methods~\cite{yue2006automated} integrating different techniques mentioned above. The deep learning technique~\cite{lecun2015deep} that emerged in recent years has achieved great success in many fields and has been widely used in medical image analysis~\cite{litjens2017survey}. It learns features with multi-level semantics automatically, which has the potential to overcome the limitations of previous methods in feature definition and extraction. Some deep learning-based methods have been proposed~\cite{arik2017fully,payer2019integrating} on this issue, but they are comparable with previous state-of-the-art methods without prominent improvement.

In this paper, we propose an end-to-end deep learning framework that can accurately and efficiently detect landmarks automatically. Our network architecture contains three sequential modules: a feature extraction module, an attentive feature pyramid fusion (AFPF) module, and a prediction module. In the feature extraction module, we use VGG-19~\cite{simonyan2014very} as a backbone network. For the critical module AFPF, we design it from two observations, while existing methods lack such considerations. One is that features extracted by different layers of neural network have various resolutions and semantics, usually higher semantics along with lower resolution. 
Identifying the landmarks on the boundary requires high-resolution and detailed structural information, while identifying the landmarks in the center of the region requires deep semantic information. To meet the requirements of identifying all the landmarks, we fuse different levels of features to get a high-resolution and semantically enhanced fusion feature. The other is that individual landmark has its specific attention to the same feature. We utilize the self-attention mechanism to learn corresponding weights of the fusion feature for different landmarks. Results show that the novel AFPF module plays an important role in improving the accuracy. It is also very flexible and can be inserted into other networks to improve the semantic representation. In the prediction module, we get inspiration from the traditional method which takes cropped patches to predict the offset of the ground truth landmarks. We adopt the combination of heat maps and offset maps to do pixel-wise regression-voting, which performs more effectively.

We evaluate the performance on the public available dataset from the ISBI Grand Challenges 2015. Our landmark detection accuracy on the validation dataset (Test Dataset 1) is 86.67\% within the clinically accepted precision range of 2.0mm with the average error of 1.17mm. As for the testing dataset (Test Dataset 2), we get the 75.05\% accuracy rate in the range of 2.0mm with the average error of 1.48mm. Our method outperforms the state-of-the-art by 7\%$\sim$11\% for all the measurements. The contributions of this paper are as follows.
\begin{itemize}
\item {Propose a new deep learning-based framework for cephalometric landmark detection.}
\item {Present a new and flexible module (AFPF) to get high-resolution and semantically enhanced fusion features with self-attention mechanism.}
\item {Our method outperforms the state-of-the-art by 7\%$\sim$11\% for all the measurements on a public dataset.}
\item {Our method has strong self-adaptive capability and performs well on unseen data source, which is very practical in clinical application.}
\end{itemize}

\begin{figure*}
 \vspace*{-5ex}
  \includegraphics[width=\textwidth]{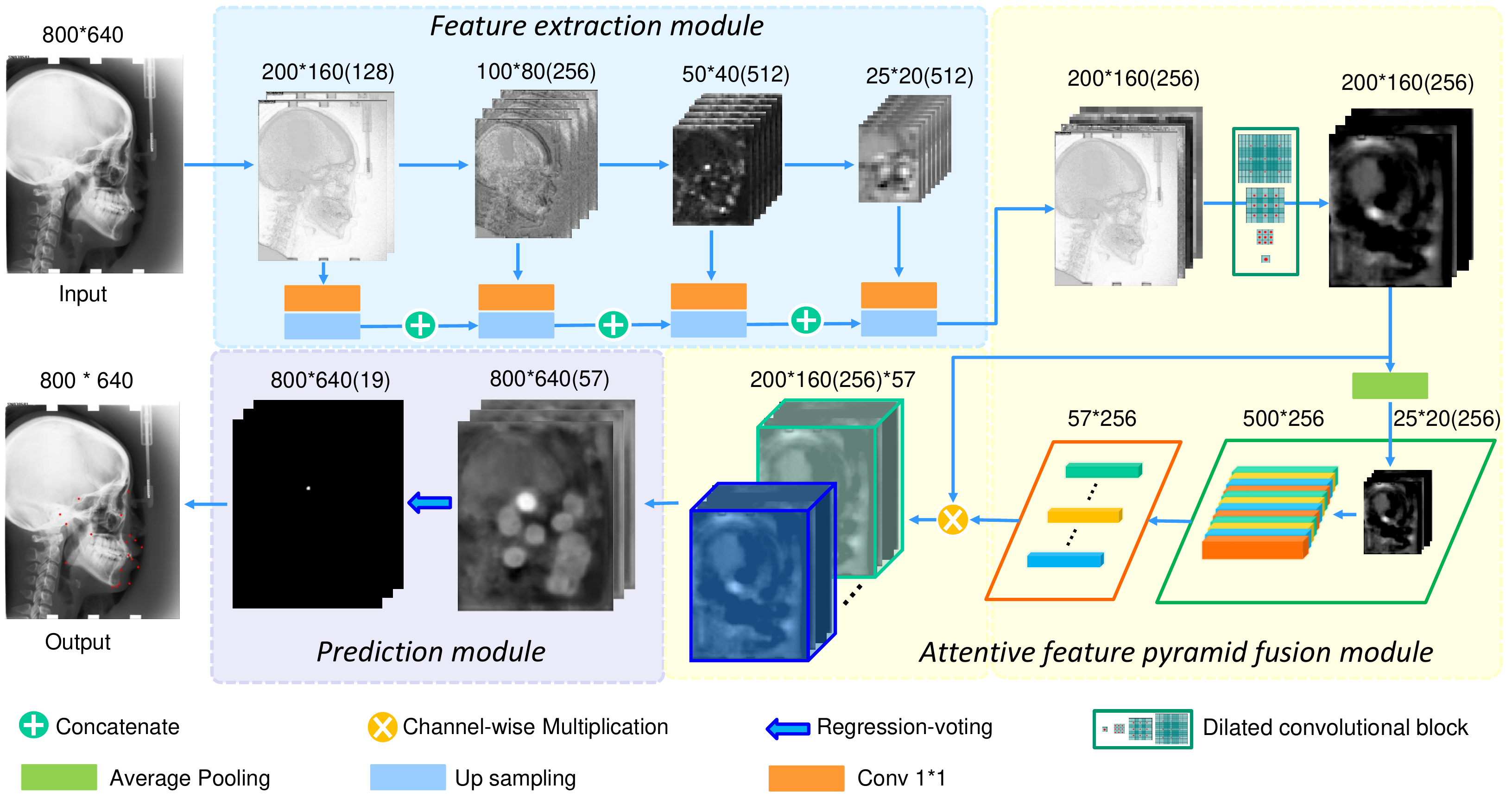}
  \vspace*{-4ex}
  \caption{Overview of our framework. Three consecutive modules, the feature extraction module, the AFPF module, and the prediction module are in blue, yellow and purple areas respectively. Feature maps are reshaped in the green box of AFPF. Attention vectors of heat maps and offset maps for landmarks are in the orange box of AFPF.}
  \label{fig:pipeline}
  \vspace*{-8ex}
\end{figure*}

\section{Methods}
Given a cephalometric radiography $I$, the goal is to detect anatomical landmark positions $P=(p_1,p_2,...,p_n)$ automatically, where $p$ denotes the 2D position for a landmark, and $n$ is the number of cephalometic landmarks. Our proposed framework is illustrated in Fig.~\ref{fig:pipeline}. In the first module, we use pre-trained VGG-19~\cite{simonyan2014very} as the backbone network. There are other networks for feature extraction like ResNet~\cite{he2016deep} and Inception~\cite{szegedy2015going}. We will show their respective performance in section 3. In the following, we give details of AFPF module and the prediction module.

\subsection {Attentive Feature Pyramid Fusion}

The Attentive Feature Pyramid Fusion (AFPF) module takes different level's features in the feature extraction module as input and produces a tensor $T$ with the size $(3n,h,w)$, where $(h,w)$ denotes the spatial size of the input image and $3n$ stand for $n$ heat maps and $2n$ offset maps. Heat maps $H$ are used to indicate the rough area of the landmark while offset maps $O$ are taken as regressors to locate the precise position~\cite{papandreou2017towards}. 
As Fig.~\ref{fig:pipeline} shows, for different levels' features in the first module, we apply $1\times1$ lateral connections and upsampling on each of them to generate feature maps with the same resolution and number of channels. Then, these feature maps are concatenated together and passed through a dilated convolutional~\cite{yu2015multi} block to form the feature pyramid $F$ with the size of $(c,h_F,w_F)$. The dilated convolution enlarges the receptive field and aggregate multi-scale context so that more local information can be used to improve the estimation accuracy. Based on the observation that different landmarks have different attention to these feature maps, we use self-attention mechanism ~\cite{vaswani2017attention,Ma2018TrafficPredict} to learn attention weights for each landmark. The attention weight $a_k$ for the $k$-th landmark is computed as follows:

\begin{equation}
a_k = \textup{softmax}({W_{k1}\tanh{(W_{k2}\tilde{F})}}),
\end{equation}
where $a_k$ is an attention matrix composed of three attention vectors $(a_k^1,a_k^2,a_k^3)$, one for heat map and two for offset maps. 
The length of each attention vector is $c$ which is equal to the channel number of $F$. $\tilde{F}$ is obtained by operations of average pooling and reshaping that transfers $F$ from the size of $(c,h_F,w_F)$ to the size $(c,h_F\times w_F/ 64)$. $W_{k1}$ and $W_{k2}$ are trainable matrices presented by fully connected layers without bias. For each landmark, we apply the attention weight $a_k$ on the feature pyramid $F$ with channel-wise multiplication to get the weighted feature pyramids $F_{k}$:

\begin{equation}
F_{k} = c(a_k \otimes F),
\end{equation}
where $F_{k}$ contains three weighted feature pyramids $(F_k^1,F_k^2,F_k^3)$, and $F_k^j ( j=1,2,3)$ has the same size as $F$. $\otimes$ is the channel-wise multiplication and $c$ is the channel number which serves as the scale factor. By applying $1\times 1$ convolution on $F_{k}$, we get the output with three channels for the $k$-th landmark, containing one heat map $H_k'$ and two offset maps $O'_k$. Then, $H_k'$ and $O'_k$ are upsampled to match the size of the input image and are named $H_k$ and $O_k$. The AFPF module can also be used in other networks to improve the semantic representation. We show its flexibility in section 3.

\subsection {Landmark Prediction with Regression Voting}

In the prediction module, we combine the output of AFPF module - heat maps and offset maps to predict landmark positions. 

In the training stage, for each pixel location $x_i$ and the $k$-th landmark $l_{k}$, we constrain the produced probability in the heat map $H_k(x_i)$ to be $1$ if $\left\|x_i-l_k\right\|_2 \leq R$ and $0$ otherwise. Here $R$ is the radius of a circular domain. Note that the heat maps and offset maps are generated by the fused feature maps of different resolution. So R is set to 40 to ensure that there is a minimal corresponding activation area on the smallest feature map (stride of 32 pixels to the input size). The loss function $L_{h}$ is defined to be mean logistic losses between the predicted heat maps and the ground truth. The offset maps are used to predict the 2D ($x$, $y$ direction respectively) offset vector $O_k(x_{i})=(l_{k}-x_{i})/R$ from the pixel $x_{i}$ to the corresponding landmark $l_{k}$. The loss function $L_{o}$ is defined to be the $L1$ loss between the predicted offsets and the target. We only calculate the loss for positions $x_{i}$ within $R$ instead of all pixels in training offset maps. The final loss function is defined as follows:

\begin{equation}\label{equ:overalllossfunc}
L(\theta) = \alpha L_{h}(\theta)+(1-\alpha)L_{o}(\theta)
\end{equation}
where $\alpha$ is a factor to balance the loss function terms. We set $\alpha = 2/3$ empirically.

In the testing stage, we aggregate the heat map and the offset maps for each landmark to construct an activation map $M_k$ via pixel-wise regression-voting as follows:

\begin{equation}
M_k(x_{i})=\sum_{x_{j}\in A_k}\mathbbm{1}\{\Vert x_{j}+\lfloor O_k(x_{j})\times R\rfloor-x_{i} \Vert=0\}
\end{equation}
where $A_k$ is the set of pixels with the $\pi R^2$ largest values in heat map $H_k$ for the $k$-th landmark, and $\mathbbm{1}\{\cdot\}$ is the indicator function. Finally, the pixel $x_{i}$ with the highest activation value $M_k(x_{i})$ is regarded as the most likely landmark position.

\section{EVALUATIONS AND DISCUSSIONS}
In this section, we first evaluate our method on the public dataset from the IEEE ISBI 2015 Challenge by comparing with state-of-the-art methods. To highlight the contribution of different parts of the framework, we also show the performance of different configurations in the ablation study. Especially, our experiments illustrate the flexibility of the AFPF module. Furthermore, we test our framework on other large-scale datasets from various devices in the extended experiments. 
The feature extraction module is pre-trained on ImageNet dataset \cite{krizhevsky2012imagenet}. We resize the input image to $800\times640$. The entire framework is built on PyTorch and optimized by the Adadelta optimizer with default configuration. The batch size is 1. The training time is approximately 7 hours for 350 epochs on a GTX 1080 TI GPU.

\subsection{Cephalometric Landmark Dataset}
The IEEE ISBI 2015 Challenge~\cite{wang2016benchmark} provides a public dataset for cephalometric landmark detections, which is the only related public dataset. The dataset consists of 400 cephalometric radiographs with 19 manually labeled landmarks by two doctors in each image, and the ground truth is the average of annotations of the two doctors. The image resolution is $1935\times2400$ pixels in the TIFF format, and the pixel spacing is 0.1mm. The pathology types for eight standard measurement methods can be calculated based on landmarks positions. We use 150 images for training, 150 images for validating and 100 images for testing, and adopt the evaluation metrics according to the IEEE ISBI 2015 Challenge standards~\cite{wang2016benchmark}, which include the mean radial error (MRE), the successful detection rate (SDR) in four target radius (2mm, 2.5mm, 3mm, 4mm), and the accuracy rate for pathology classification (APC).

\begin{table*}[h]
	\centering
	\caption{Comparison with three state-of-the-art methods and ablation study of our own method based on the dataset of IEEE ISBI 2015 Challenge. The model index from 1 to 6 stands for Ibragimov et al.~\cite{ibragimov2015computerized}, Lindner et al.~\cite{lindner2015fully}, Arik et al.~\cite{arik2017fully}, our method without AFPF module, our method without self-attention mechanism, and our complete method respectively. }\label{tab:tab2}
	\begin{tabular}{c|c c c c c c|c c c c c c}
		\hline
		\multirow{2}*{\tabincell{c}{Model}} & \multicolumn{6}{c|}{Test dataset 1} & \multicolumn{6}{c}{Test dataset 2}\\
        \cline{2-13}
		~ & MRE& 2mm & 2.5mm & 3mm & 4mm & APC & MRE & 2mm & 2.5mm & 3mm & 4mm & APC\\
		\cline{1-13}
		1& 1.84 & 71.72 & 77.40& 81.93& 88.04 & 70.84&- & 62.74& 70.47& 76.53& 85.11 & 76.12\\
		2 & 1.67 & 73.68& 80.21& 85.19& 91.47 & 76.41&- & 66.11& 72& 77.63& 87.42 & 80.99\\
		3 &- & 75.37& 80.91& 84.32& 88.25 & 75.92 & - & 67.68& 74.16& 79.11& 84.63 & 76.75\\
		4 & 1.35&83.96 &90.63 &94.00 &97.64 & 77.50 &1.55&73.58 &81.95 &87.68 &94.31  & 81.41\\
		5  & 1.22&85.64&91.89&95.09&98.35 & 78.74&1.52&74.21&82.11&87.47&	94.74 & 80.61\\
		6 & \textbf{1.17}&\textbf{86.67} &\textbf{92.67} &\textbf{95.54} &\textbf{98.53} & \textbf{79.05} &\textbf{1.48} &\textbf{75.05} &\textbf{82.84} &\textbf{88.53} &\textbf{95.05}  & \textbf{81.95}\\
		
		\hline
	\end{tabular}
	\label{table1}
\end{table*}

\begin{table*}[h]
	\centering
	\caption{Performance of the AFPF module on other feature-extraction networks. Results by removing AFPF are shown in each second line.}
	\begin{tabular}{c|c c c c c|c c c c c}
		\hline
		\multirow{2}*{\tabincell{c}{Model}} & \multicolumn{5}{c|}{Test dataset 1} & \multicolumn{5}{c}{Test dataset 2}\\
        \cline{2-11}
		~ & MRE& 2mm & 2.5mm & 3mm & 4mm & MRE & 2mm & 2.5mm & 3mm & 4mm \\
		\cline{1-11}
		ResNet50 & \textbf{1.18}&\textbf{86.14} &\textbf{91.72} &\textbf{95.79} &\textbf{98.25} &\textbf{1.52} &\textbf{73.37} &\textbf{81.37} &\textbf{87.53} &\textbf{94.73}  \\
		ResNet50\_noAFPF &
		1.40&80.74 &88.84 &93.19 &97.51 &1.63&70.05 &80.63 &86.21 &94.26 
		 \\
		\cline{1-11}
		Inception & \textbf{1.25}&\textbf{86.21} &\textbf{91.89} &\textbf{95.65} &\textbf{98.18} &\textbf{1.47} &\textbf{73.89} &\textbf{83.37} &\textbf{88.26} &\textbf{95.26}  
		 \\
		Inception\_noAFPF & 1.50&78.53 &86.84 &91.36 &96.11 &1.79&66.95 &76.00 &84.00 &93.11  \\
		\hline
	\end{tabular}
	\label{table2}
	\vspace*{-2ex}
\end{table*}

\subsection{Baselines}
We compare our approach with the top two methods~\cite{lindner2015fully,ibragimov2015computerized} in IEEE ISBI 2015 Challenge and two new approaches proposed by Arik et al.~\cite{arik2017fully} and Payer et al.~\cite{payer2019integrating}. We also remove the AFPF module and the attention mechanism respectively to do the ablation study.

\subsection{Analysis}
The comparison with state-of-the-art methods is shown in Table~\ref{table1}. Our method (results are marked by black body) has much better performance under all the measurements, achieving the MRE of 1.17mm and 1.48mm with the standard deviation of 1.19mm and 0.77mm for the test dataset 1 and test dataset 2 respectively. In terms of the successful detection rate (SDR) for target radius evaluated in the Challenge, our method is higher than the top method by 7\%$\sim$11\%. As for the accuracy rate for pathology classification (APC), we achieved 79.05\% and 81.95\% respectively on two test datasets, with an average classification accuracy of 84.7\% for all classified subjects. Payer et al.~\cite{payer2019integrating} show results under the other format. They count accuracy rates in four target radius by combining two test datasets and get 73.33\%, 78.76\%, 83.24\%, and 89.75\% respective while we get much higher accuracy of 82.03\%, 88.74\%, 92.74\%, and 97.14\%. The ablation study in Table~\ref{table1} shows that the AFPF module really helps our model improve the detection accuracy and the attention mechanism plays an important role in the AFPF module. Note that even without the AFPF module, our network, adopting pixel-level regression-voting technique based on heat maps and offset maps, has already outperformed others.  

To further identify the flexibility of the AFPF module, we replace the VGG-19 in the first module of our framework by other feature-extraction networks, including ResNet50 and Inception. The results in Table~\ref{table2} illustrate that our AFPF module can be used on many networks to improve the performance. In addition, our method is efficient, processing one image within 70ms on a GTX 1080 TI GPU or 7.8s on the i7-6700K CPU.

\begin{figure*}
  \vspace*{-4ex}
  \includegraphics[width=\textwidth]{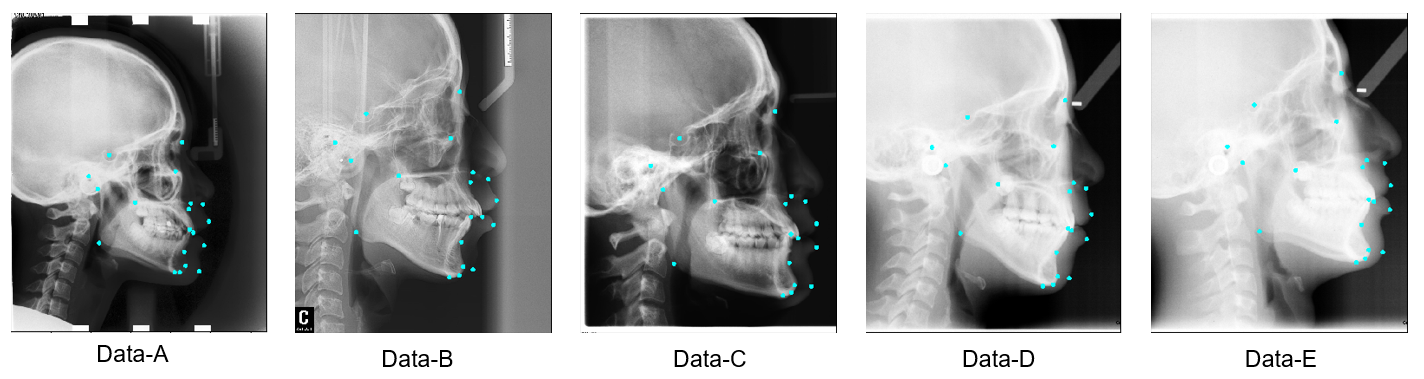}
  \vspace*{-5ex}
  \caption{Data samples of five datasets with 393, 154, 100, 709, 501 samples respectively. Blue points are the ground truth of landmarks.}
  \label{fig:samples}
  \vspace*{-8ex}
\end{figure*}

\subsection{Extended Experiments}
In the real world, images captured by different devices differs significantly. A good algorithm should have the self-adaptive ability even for the data from a new device. To test generalization capability and stability of our approach, we use five datasets (Fig.~\ref{fig:samples}) collected by four different devices. We name them as Data-A, Data-B, Data-C, Data-D, and Data-E, where Data-D and Data-E are from the same device. All of them are manually relabeled 19 landmarks by a dentist to avoid inter-observer errors. We use the Data-A, Data-C, and Data-D for training while Data-B and Data-E for testing. 
For the testing Data-E, the MSE is 1.03mm with 94.2\%, 96.86\%, 98.16\% and 99.31\% SDRs in the measurement of 2mm, 2.5mm, 3mm, 4mm respectively. For Data-B, which comes from a new device and never occurs in the training dataset, the MSE is 0.88mm with 94.73\%, 97.56\%, 98.80\% and 99.66\% SDRs. It shows that our method is well performed for seen or unseen data sources and very practical for the clinical application.

\section{Conclusion}

We propose an end-to-end deep learning framework to automatically detect cephalometric landmarks with high accuracy. In our framework, the AFPF module can get high-resolution and semantically enhanced fusion feature with attention to improve the prediction accuracy. The pixel-wise regression-voting technique based on heat maps and offset maps also benefits the performance. Our framework achieves state-of-the-art performance for all the evaluation metrics. Especially, our method performs well even for unseen data sources, which is significant for the actual application. Our framework is able to take in raw images and output landmarks directly in real time without any human intervention, so it is useful for fully automated cephalometric analysis. In the future, we will extend our method to more general landmark-detection tasks.

\section*{Acknowledgment}
This work is supported partially by Innovative Technology Fund (ITS/411/17FX) and General Research Fund (17210419), Hong Kong SAR.

\end{document}